\documentclass[letterpaper]{article} 
\usepackage{aaai25}  
\usepackage{times}  
\usepackage{helvet}  
\usepackage{courier}  
\usepackage[hyphens]{url}  
\usepackage{graphicx} 
\usepackage{natbib}  
\usepackage{caption} 
\usepackage{algorithm}
\usepackage{algorithmic}
\usepackage{amsmath}
\usepackage{amssymb}
\usepackage{pifont}
\usepackage{makecell}
\usepackage{colortbl}
\usepackage[table]{xcolor}
\usepackage{xspace}
\usepackage{url}
\usepackage[outdir=./picture/]{epstopdf}

\newcommand{\yes}{\ding{51}}
\newcommand{\no} -
\newcommand{\dataname}[0]{EMHI\xspace}
\newcommand{\numseq}{885\xspace}
\newcommand{\numpeople}{58\xspace}
\newcommand{\numtime}{28.5\xspace}
\newcommand{\numaction}{39\xspace}
\newcommand{\numpair}{3.07M\xspace}
\usepackage{newfloat}
\usepackage{listings}
\DeclareCaptionStyle{ruled}{labelfont=normalfont,labelsep=colon,strut=off} 
\floatstyle{ruled}
\newfloat{listing}{tb}{lst}{}
\floatname{listing}{Listing}
\title{EMHI: A Multimodal Egocentric Human Motion Dataset with HMD and Body-Worn IMUs}
\author{
    Zhen Fan\equalcontrib, Peng Dai\equalcontrib, Zhuo Su\equalcontrib, Xu Gao, Zheng Lv, Jiarui Zhang, \\Tianyuan Du, Guidong Wang, Yang Zhang\\ 
}
\affiliations{
    PICO\\
    \{fanzhen.0315, daipeng.2022, suzhuo, gaoxu.1024, lvzheng.101, zhangjiarui.zjr123,\\dutianyuan, guidong.wang, zhangyang.0621\}@bytedance.com
}

\usepackage{bibentry}
\usepackage{booktabs}
\usepackage{multirow}

\begin{document}

\maketitle

\begin{abstract}
Egocentric human pose estimation (HPE) using wearable sensors is essential for VR/AR applications. Most methods rely solely on either egocentric-view images or sparse Inertial Measurement Unit (IMU) signals, leading to inaccuracies due to self-occlusion in images or the sparseness and drift of inertial sensors. Most importantly, the lack of real-world datasets containing both modalities is a major obstacle to progress in this field. To overcome the barrier, we propose \dataname, a multimodal \textbf{E}gocentric human \textbf{M}otion dataset with \textbf{H}ead-Mounted Display (HMD) and body-worn \textbf{I}MUs, with all data collected under the real VR product suite. Specifically, \dataname provides synchronized stereo images from downward-sloping cameras on the headset and IMU data from body-worn sensors, along with pose annotations in SMPL format. This dataset consists of \numseq sequences captured by \numpeople subjects performing \numaction actions, totaling about \numtime hours of recording. We evaluate the annotations by comparing them with optical marker-based SMPL fitting results. To substantiate the reliability of our dataset, we introduce MEPoser, a new baseline method for multimodal egocentric HPE, which employs a multimodal fusion encoder, temporal feature encoder, and MLP-based regression heads. The experiments on \dataname show that MEPoser outperforms existing single-modal methods and demonstrates the value of our dataset in solving the problem of egocentric HPE. We believe the release of \dataname and the method could advance the research of egocentric HPE and expedite the practical implementation of this technology in VR/AR products. Project page at: \textcolor{red}{\textit{\url{https://pico-ai-team.github.io/EMHI/}}}
\end{abstract}

\section{Introduction}
\label{introduction}

\begin{figure*}[!htb]
  \centering
  \includegraphics[width=0.97\linewidth]{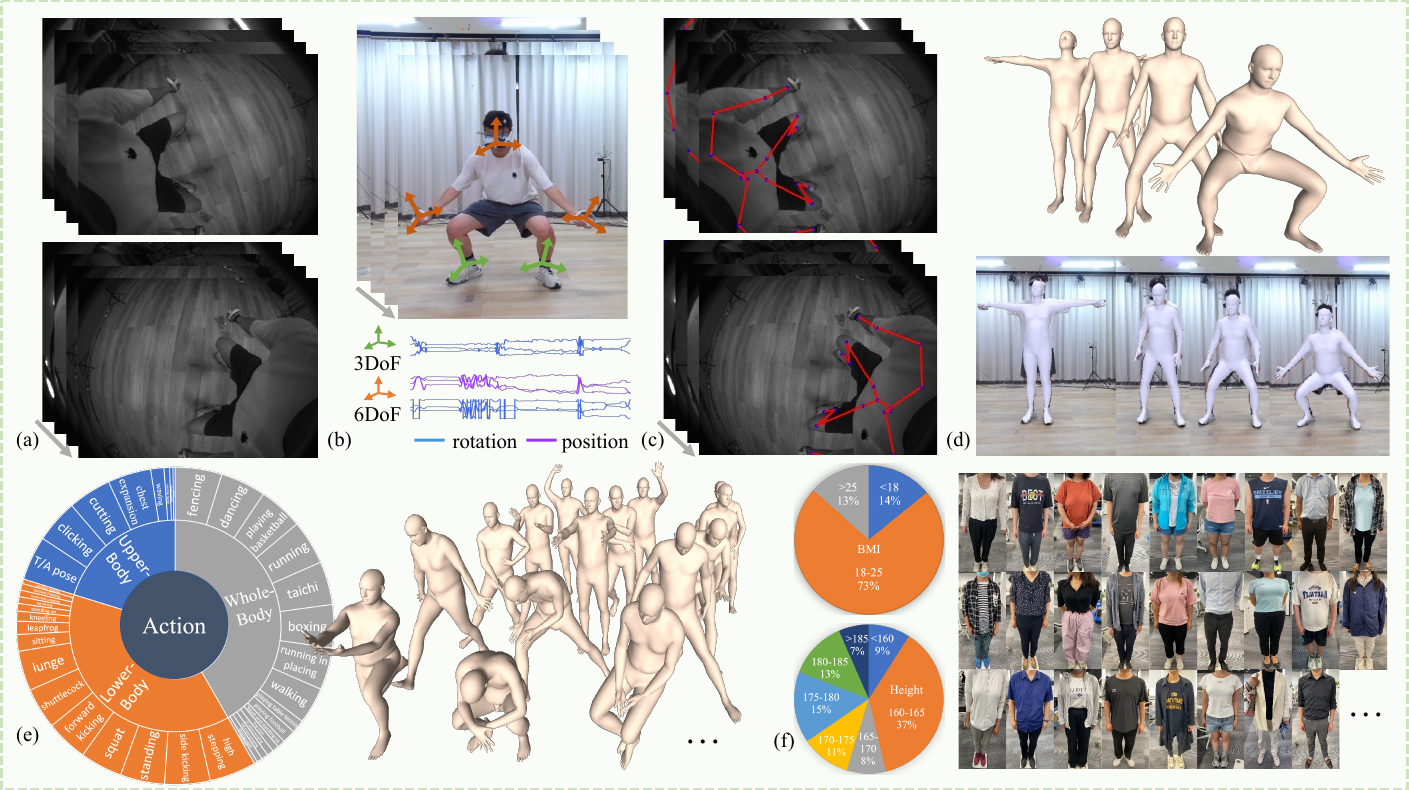}
  \caption{\dataname is a multimodal dataset that provides (a) stereo egocentric images and (b) IMU signals.The annotations include (c) 2D keypoints overlaid on egocentric images and (d) SMPL parameters in the world coordination. Each sequence is also annotated with (e) the action label, as well as (f) individual attributes such as height, BMI, and clothing descriptions.}
  \label{fig-overview}
\end{figure*}

Egocentric human pose estimation (HPE) has gained significant attention in computer vision, driven by the demand for accurate motion tracking in immersive VR/AR environments. Unlike traditional exocentric HPE which relies on external sensors, egocentric HPE employs body-worn sensors such as egocentric cameras or sparse IMUs. Although rapid progress has been made in this field, there remain challenges in obtaining accurate full-body poses from single-modal data due to issues like 1) self-occlusion and viewpoint variations in egocentric vision; and 2) sparsity and drifting of IMU data. Most importantly, the lack of real-world multimodal training data poses the most significant challenge.

Previous works~\cite{rhodin2016egocap,xu2019mo} introduced egocentric datasets using experimental fisheye camera setups to capture images and annotate 3D joints. However, these setups are impractical for real VR/AR products, which need compact, lightweight designs. Synthetic datasets~\cite{tome2019xr,akada20243d,cuevas2024simpleego} use physics engines for egocentric image rendering but suffer a domain gap with real images due to the complexity of human motion and environments. Meanwhile, the lower body may be occluded in a vision-based setting, and some body parts may fall outside the field of view (FOV) depending on the body pose. IMU-based datasets avoid occlusion but suffer from drift over time and ill-posed problems from sparse observations. Besides, existing methods~\cite{jiang2022avatarposer,zheng2023realistic, dai2024hmd} typically use synthetic IMU data from AMASS~\cite{mahmood2019amass}, which may not accurately reflect real-world noise and drift. Some datasets~\cite{trumble2017total, huang2018deep} provide real 3 Degrees of Freedom~(3DoF, rotation) data from XSens, while others~\cite{dai2024hmd} include 6DoF~(rotation and position) data of head, hands, and 3DoF for lower legs, but these are small-scale and primarily used for evaluation. Recently, several large-scale multimodal datasets~\cite{ma2024nymeria,grauman2024ego} have been released, offering RGB images, upper body IMU data, and motion narrations. However, the forward-facing camera limits the egocentric view, and missing lower-body IMU signals can cause ambiguity.

Combining egocentric cameras and body-worn IMUs offers a promising multimodal solution due to their lightweight and flexible design. This configuration is also commonly found in VR scenarios. Our proposed \dataname dataset, as shown in Fig. \ref{fig-overview}, features a VR headset with two downward-sloping cameras for egocentric image capture, 6DoF head and hand tracking, and additional IMUs on an actual VR device for lower-leg 3DoF tracking. We use a markerless multi-view camera system for SMPL~\cite{SMPL:2015} ground truth acquirement, with accuracy and consistency refinement using IMU data, and synchronization via OptiTrack. Furthermore, we propose a new baseline method, MEPoser, integrating egocentric images and IMU data to perform real-time HPE on a standalone VR headset. The method employs a multimodal fusion encoder, a temporal feature encoder, and MLP-based regression heads to estimate SMPL body model parameters, effectively demonstrating the advantages of multimodal data fusion in enhancing pose accuracy and the value of our dataset. This approach paves the way for further research in egocentric HPE using multimodal inputs.

In summary, our work makes the following contributions:
\begin{itemize}
\item We first introduce a large-scale multimodal egocentric motion dataset \dataname on the real VR device, including stereo downward-sloping egocentric images, full-body IMU signals, and accurate human pose annotations.
\item We propose a baseline method MEPoser, which employs a multimodal fusion encoder, temporal feature encoder, and MLP-based regression heads to perform real-time HPE on a standalone HMD. 
\item The experiment results demonstrate the rationality of our multimodal setting and the effectiveness of \dataname for addressing egocentric HPE.
\end{itemize}

\section{Related Work}
\label{relatedwork}

\begin{table*}[!t]
  \centering
  \renewcommand{\arraystretch}{0.6}
  \rowcolors{4}{gray!25}{white}
  \begin{tabular}{m{1.6cm}<{\centering}m{1cm}<{\centering}m{1.5cm}<{\centering}m{2.6cm}<{\centering}m{2.6cm}<{\centering}m{1.5cm}<{\centering}m{1cm}<{\centering}m{1cm}<{\centering}m{1cm}<{\centering}} 
    \toprule
    \multirow{2}{*}{Dataset} & \multirow{2}{*}{Device}  & \multirow{2}{*}{Real/Synth} & \multicolumn{2}{c}{Sensor Modality} & \multirow{2}{*}{SMPL(x)}   & \multicolumn{3}{c}{Statistic}\\
    \cmidrule{4-5} \cmidrule{7-9}
    &&&Egocentric Vision&Inertial&&Actions & Subjects & Frames\\
    \midrule
    $Mo^2Cap^2$ & \includegraphics[width=0.05\textwidth,height=0.05\textwidth]{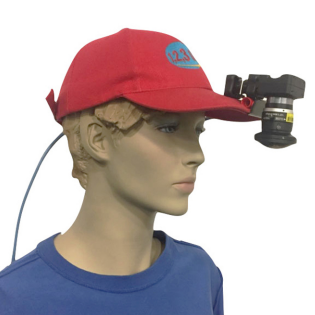} &Synth. & \makecell[c]{Monocular \\ Downward-Facing} & \no & \no   & 3K &700& 530K \\
    
    EgoPW& \includegraphics[width=0.05\textwidth]{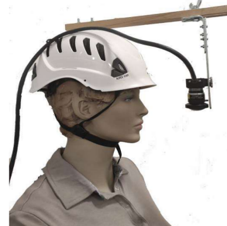} & Real. & \makecell[c]{Monocular \\ Downward-Facing} & \no & \no   & 20 & 10 &  318K \\
    
    EgoCap & \includegraphics[width=0.05\textwidth, height=0.05\textwidth]{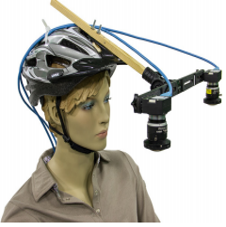} & Real & \makecell[c]{Binocular \\ Downward-Facing} & \no  &  \no   & - & 8& 30K \\
    
    UnrealEgo & \includegraphics[width=0.05\textwidth, height=0.05\textwidth]{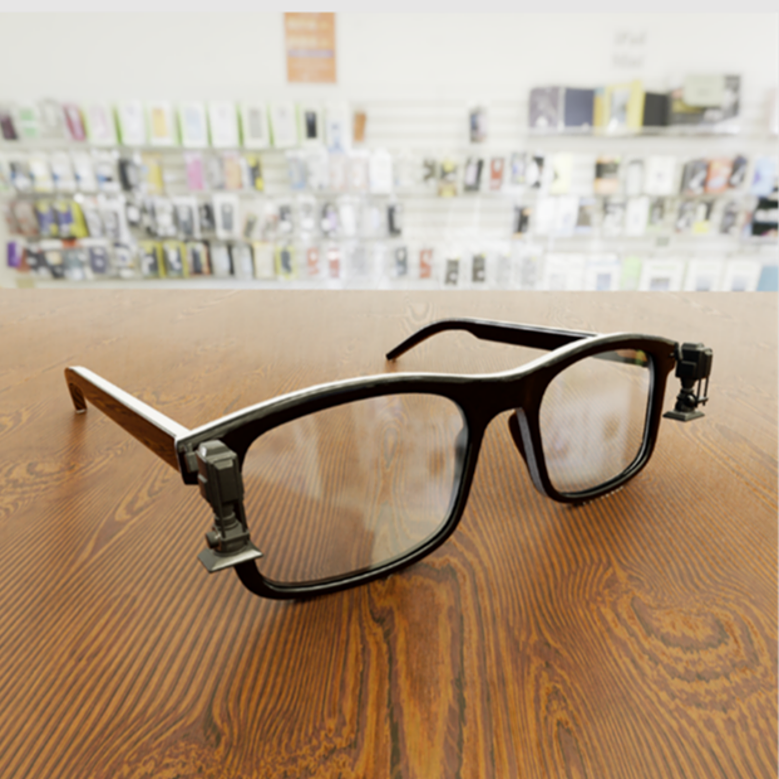}& Synth. & \makecell[c]{Binocular \\ Downward-Facing} & \no & \no   & 30 &17& 450K \\
    
    DIP-IMU & \includegraphics[width=0.05\textwidth, height=0.05\textwidth]{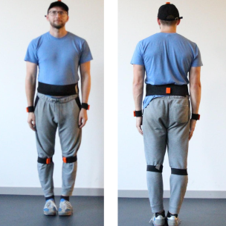}& Real & \no & \makecell[c]{Full-Body \\ 3DoF$\times$6 }  & \yes   & 15 &10 & 330K \\
    
    FreeDancing &\includegraphics[width=0.05\textwidth, height=0.05\textwidth]{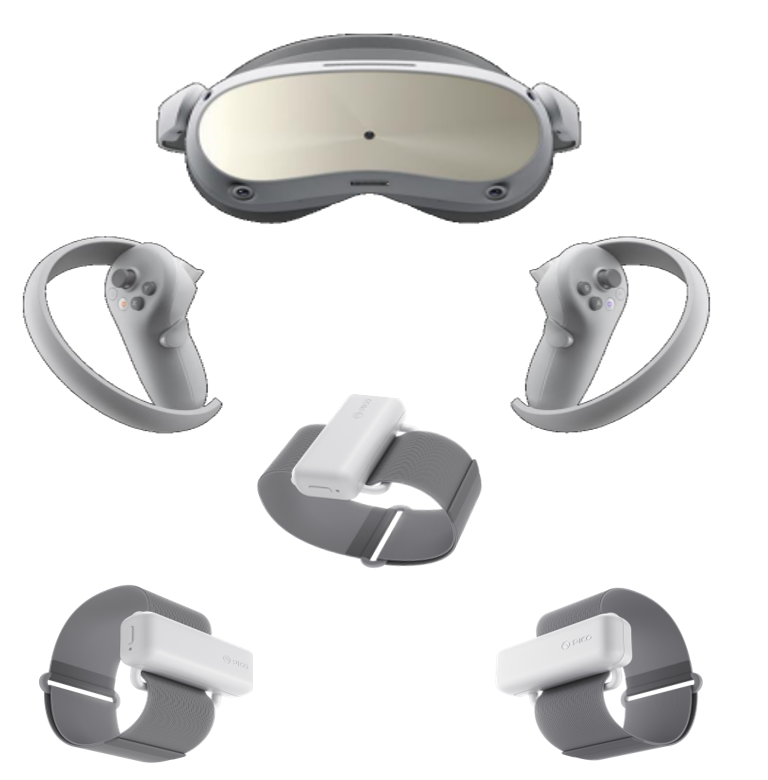}  & Real & \no &\makecell[c]{Full-Body \\ 6DoF$\times$3, 3DoF$\times$3} & \yes  & - & 8 & 532.8K\\ 
    
    Nymeria & \includegraphics[width=0.05\textwidth, height=0.05\textwidth]{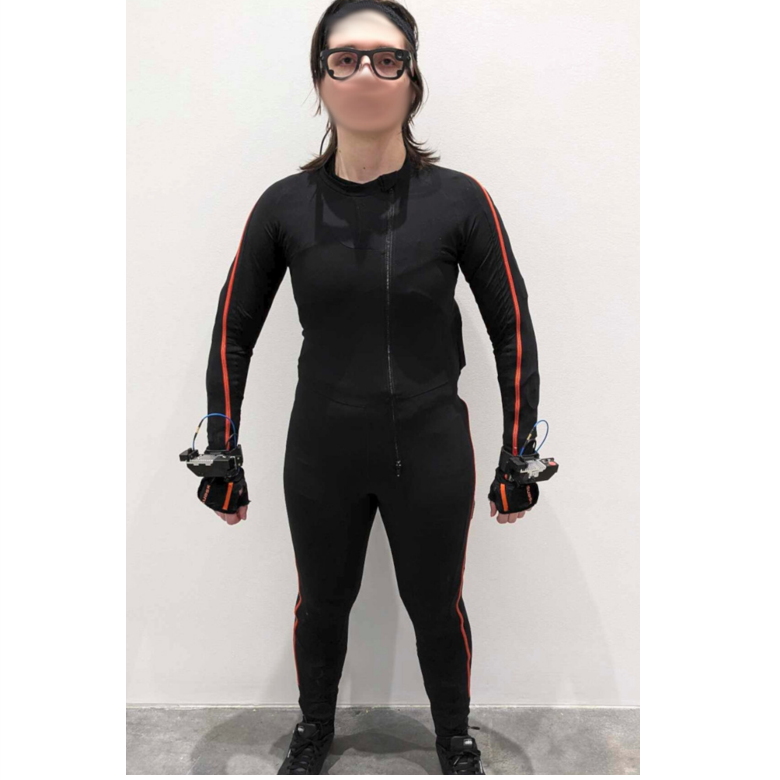}& Real &  \makecell[c]{Binocular \\ Forward-Facing} & \makecell[c]{Upper-Body \\ 6DoF$\times$3} & \yes  & 20 & 264 & 260M \\ 
    
    \makecell[c]{Ego-Exo4D\\(Ego Pose)} & \includegraphics[width=0.05\textwidth, height=0.05\textwidth]{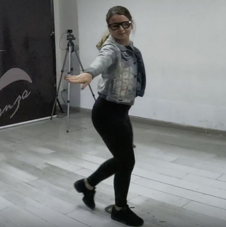} & Real & \makecell[c]{Binocular \\ Forward-Facing}  &\makecell[c]{Head \\ 6DoF$\times$1} & \no   & - & - & 9.6M \\
    
    \midrule
    \textbf{Ours} & \includegraphics[width=0.05\textwidth, height=0.05\textwidth]{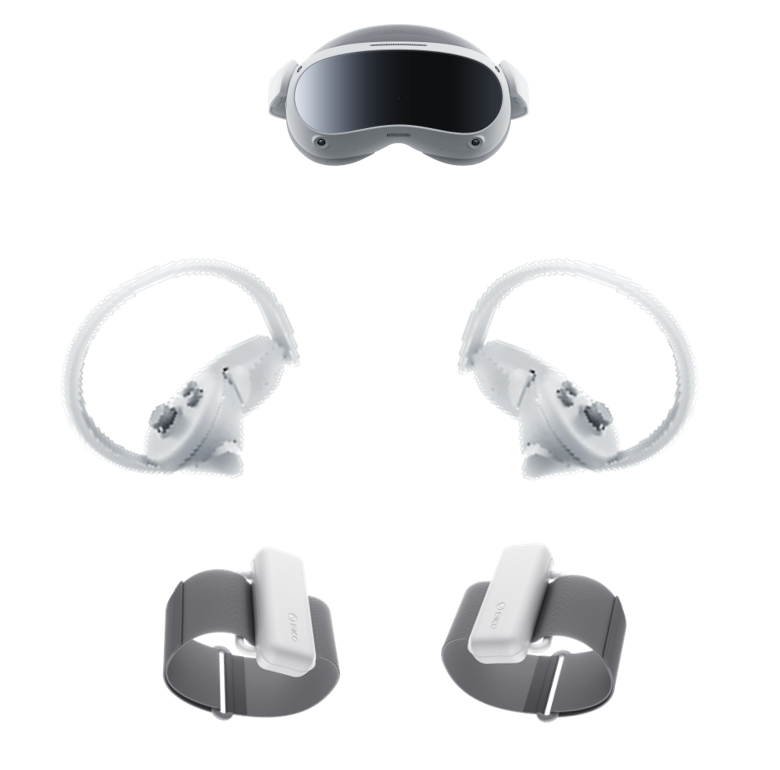} & Real & \makecell[c]{Binocular \\ Downward-Sloping} & \makecell[c]{Full-Body \\ 6DoF$\times$3, 3DoF$\times$2} & \yes  & 78 & 305 & 32.4M \\
    \bottomrule
  \end{tabular}
  \caption{Comparison with existing egocentric motion datasets. \dataname is the first dataset that provides egocentric vision and full-body IMU signals captured by the real VR product suite, along with accurate SMPL annotations simultaneously.}
  \label{tab:dataset_compare}
\end{table*}

\subsection{Egocentric Motion Dataset}
As shown in Tab. \ref{tab:dataset_compare}, existing egocentric motion datasets can be divided into vision-based, IMU-based, and multimodal datasets depending on the input modality.

Vision-based egocentric motion datasets~\cite{zhao2021egoglass,wang2022estimating,wang2023scene,liuegofish3d} provide first-person perspective images using head-mounted cameras, with the corresponding annotations of the wearer's poses. Mo2Cap2~\cite{xu2019mo} and xR-EgoPose~\cite{tome2019xr} made an early effort to build the synthetic monocular dataset with a downward-facing fisheye camera. 
EgoWholeBody~\cite{wang2024egocentric} is the latest synthetic dataset providing high-quality images and SMPL-X annotations. EgoCap~\cite{rhodin2016egocap} is a pioneer binocular dataset captured by helmet-mounted stereo cameras, containing 30K frames recorded in a lab environment. 
To relieve the dataset scale limitation, UnrealEgo~\cite{akada2022unrealego} proposed a large-scale and highly realistic stereo synthetic dataset with 450K stereo views and was extended to 1.25M in UnrealEgo2~\cite{akada20243d}. SynthEgo~\cite{cuevas2024simpleego} extended synthetic datasets with more identities and environments, annotated with SMPL-H for better body shape descriptions. 

Sparse IMU-based datasets provide an alternative for this problem. AMASS~\cite{mahmood2019amass} can provide a large-scale synthetic IMU dataset. TotalCapture~\cite{trumble2017total} and DIP-IMU~\cite{huang2018deep} offered real IMU data captured by Xsens and SMPL pose annotations obtained by marker-based optical mocap system and IMU-based method~\cite{von2017sparse} respectively. PICO-FreeDancing~\cite{dai2024hmd} provided sparse IMU data with SMPL format GT fitting using OptiTrack data. 

Multimodal datasets\cite{damen2022rescaling, gong2023mmg,cha2021mobile,rai2021home} have attracted significant attention in recent years due to the complementarity of different data modalities. 
Ego-Exo4D~\cite{grauman2024ego}, Nymeria~\cite{ma2024nymeria} and SimXR~\cite{luo2024real} captured real-world images by Project Aria glasses~\cite{somasundaram2023project}, along with the IMU data in upper-body. Ego-Exo4D provided up to 9.6M image frames with annotations of the body and hand joint positions. Nymeria further offered SMPL format data derived from Xsens mocap suits, with limited clothing diversity of the captured body. 
However, in these datasets, either the forward-facing perspective restricts the perception range of the wearer’s body, or they have not integrated downward-sloping perspectives and sparse full-body IMU signals on actual VR/AR devices.

\begin{figure*}[!tb]
    \centering    
    \includegraphics[width=0.99\linewidth]{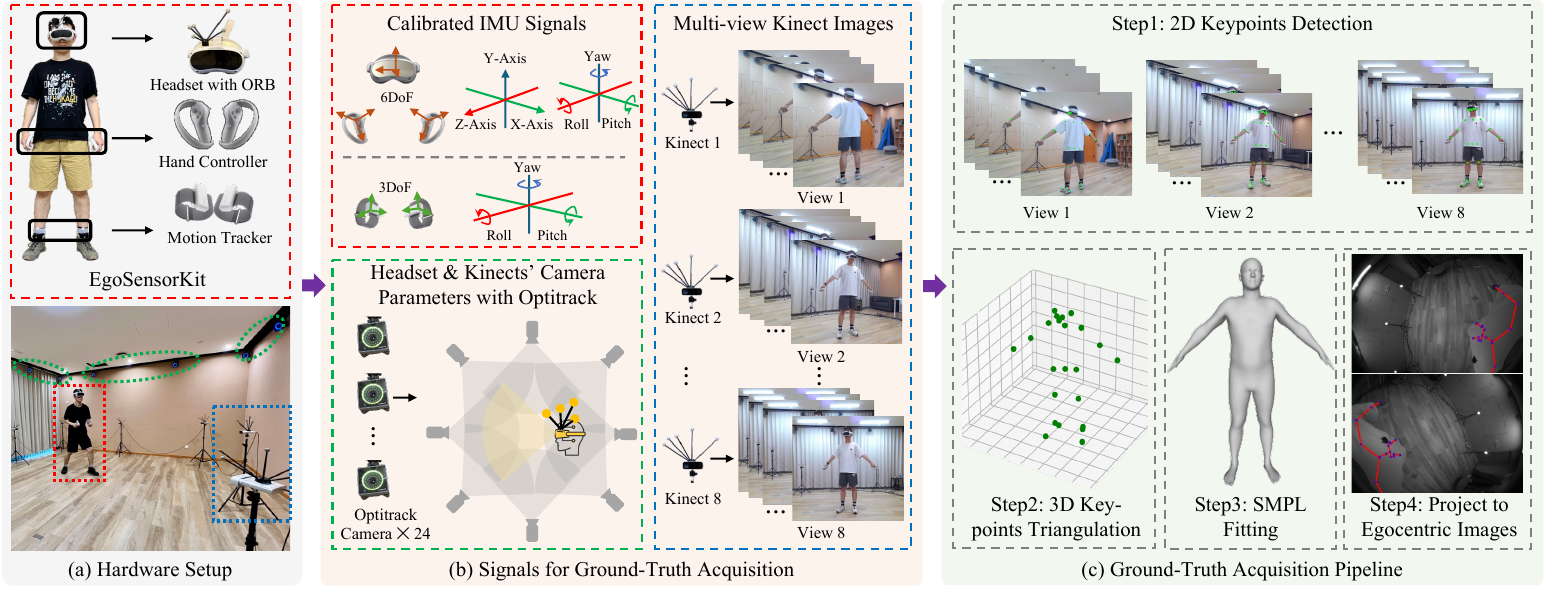}
    \caption{Hardware setup and ground-truth acquisition pipeline. (a) the data capture system consists of EgoSenorKit for egocentric images and calibrated IMU signals collection, eight Azure Kinects for multiple third-view image recording and an Optitrack system for spatiotemporal synchronization of the above signals. With the data collected in (b), (c) we produce the annotations including SMPL parameters and 2D keypoints on egocentric images automatically.}
    \label{fig-hardware}
\end{figure*}

\subsection{Egocentric Human Pose Estimation Methods}
Vision-based methods have been widely investigated ~\cite{rhodin2016egocap,Wang_2021_ICCV,wang2022estimating,wang2023scene,akada2022unrealego,liu2022ego+,liuegofish3d,kang2023ego3dpose,park2023domain,cuevas2024simpleego}.
For single egocentric pose estimation, Wang \textit{et al.}~\cite{wang2024egocentric} proposed an egocentric motion capture method that combines the vision transformer for undistorted image patch feature extracting and uses diffusion-based motion priors for pose refinement. However, 3D pose estimation from a single image remains challenging due to the lack of depth information. To address this, EgoPoseFormer~\cite{yang2024egoposeformer} introduced stereo egocentric skeleton tracking methods by a two-stage transformer-based corse-to-fine pose estimator for 3D pose prediction. Hiroyasu Akada \textit{et al.}~\cite{akada20243d} further enhanced this with a transformer-based model utilizing 3D scene information and temporal features. Despite these advancements, challenges with invisible body parts due to self-occlusion and out-of-view joints persist.

Methods using sparse tracking signals from body-worn IMUs have garnered significant attentions~\cite{winkler2022questsim,SIP,yi2021transpose,yi2022physical,TIP,AGROL}.
In egocentric VR and AR scenarios, there are inherently three 6DoF tracking points for the head and hands, with the option to add two additional 3DoF IMUs on the legs.
AvatarPoser~\cite{jiang2022avatarposer} proposed a global pose prediction framework combining transformer structures with inverse kinematics (IK) optimization, while AvatarJLM~\cite{zheng2023realistic} introduced a two-stage approach that models joint-level features and uses them as spatiotemporal transformer tokens to achieve smooth action capture. 
HMD-Poser~\cite{dai2024hmd} integrated these inputs, presenting a lightweight temporal-spatial learning method for full-body global 6DoF body action recovery.
However, IMU-based data faces challenges such as drift and sparsity. 

\section{\dataname Dataset}
\label{dataset}
\dataname is a multimodal egocentric motion dataset that contains \numpair synchronized data pairs organized as \numseq sequences recording at 30FPS. Each data pair contains stereo egocentric images (640 $\times$ 480), five IMUs data, and corresponding 3D SMPL pose and 2D keypoints. It is captured by \numpeople subjects, which are equally split into 29 male and 29 female, with a diverse range of body shapes. Each subject wears their daily clothing during data collection to ensure a wide variety of natural looks. We record \numaction common actions of users experiencing games and social applications in VR scenarios and categorize them into upper-body motions, lower-body motions, and full-body motions. Additionally, this dataset is captured under three different environmental lighting conditions: dim light, natural light, and bright light for environment diversity. 

\subsection{Data Capture System}
\subsubsection{Hardware}
\label{sec:hardware}
As shown in Fig. \ref{fig-hardware}, the overall hardware consists of three subsystems: EgoSensorKit system to collect sensor data, with a PICO4 headset, two hand controllers, and two leg trackers; Kinect system to obtain SMPL annotations, with 8 cameras recording simultaneously from outside-in viewpoint; Optitrack system for spatiotemporal synchronization between the above two systems, with Optical Rigid Body (ORBs) mounted at the VR headset and all Kinect cameras, allowing all camera moving.

\subsubsection{Temporal Synchronization}
\label{sec:time-sync}
Kinect and Optitrack systems rely on a signal transmitter device to trigger simultaneously, ensuring their inter-frame alignment. EgoSensorKit and Optitrack could also be synchronized offline with the headset IMU's and ORB's angular velocity, according to the motion correlation method in~\cite{qiu2020real}. Finally, the data frames of Kinect and EgoSensorKit are aligned via Optitrack as a bridge. As the recorded frame rate is 30Hz, the maximum synchronization deviation will reach up to 16.5 ms which might be notable during fast motion. So the annotations are further post-processed with linear interpolation to better align with the EgoSensorKit's timestamps.

\subsubsection{Spatial Alignment}
\label{sec:spatial-align}
The 6DoF of the headset (IMU) in Optitrack coordination $T_{h}^{o}$ could be obtained by $T_{h}^{o} = T_{rb}^{o}\ T_{h}^{rb}$, where $T_{h}^{rb}$ is a pre-calibrated rigid transformation between the IMU sensor in headset and its ORB, and the ORB's 6DoF $T_{rb}^{o}$ is tracked with Optitrack. Similarly, the extrinsic parameters of each Kinect RGB camera in the Optitrack coordinate system $ T_{k}^{o}$ can be determined using the same method. Then, the spatial transformation between Kinect cameras and headset could be obtained by $T_{k}^{h} = ({T_{h}^{o}})^{-1}\ T_{k}^{o}$. Finally, the transformation matrix between the Kinects and egocentric cameras could be further calculated by $T_{k}^{c} = T_{h}^{c}\ T_{k}^{h}$, where $T_{h}^{c}$ is also a constant spatial relationship between the headset (IMU) and its egocentric cameras.

\begin{figure*}[!htb]
    \centering
    \includegraphics[width=0.99\linewidth]{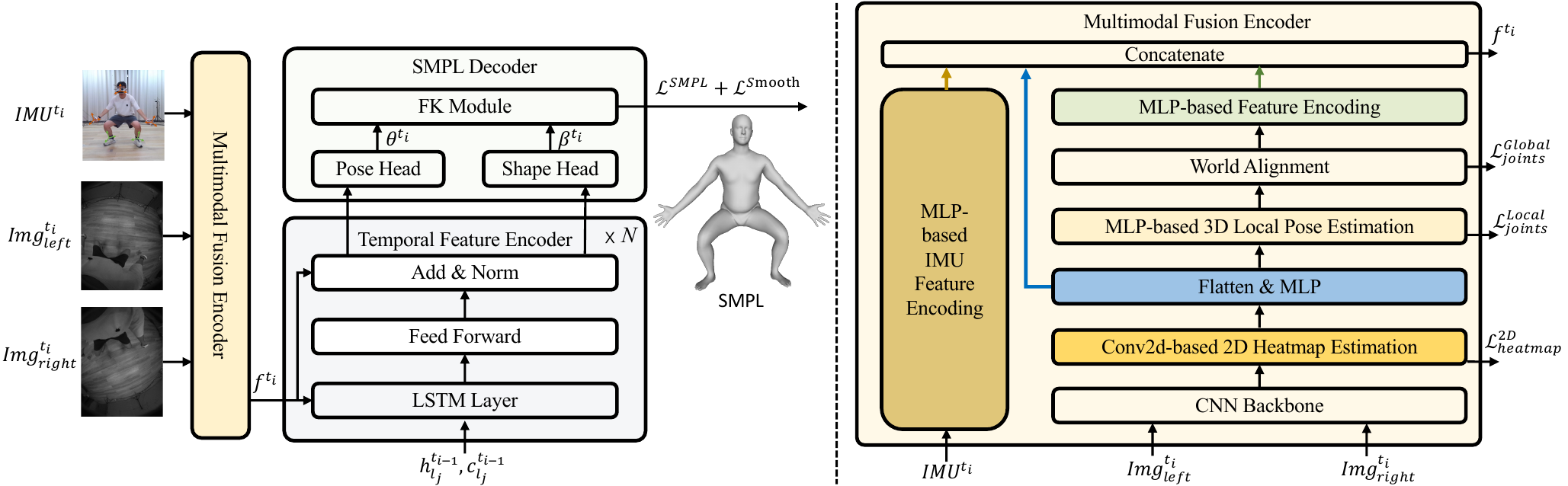}
    \caption{MEPoser. The proposed method consists of a multimodal fusion encoder for feature extracting and fusion of input signals, a temporal feature encoder for history information association, and an SMPL decoder for SMPL parameters prediction.}
    \label{fig-baseline}
\end{figure*}

\subsection{Ground-Truth Acquisition}
\label{annotation}
\subsubsection{Keypoints Annotation}
We use HRNet~\cite{sun2019deep} to detect 2D keypoints in multi-view Kinect RGB images with the body25 format~\cite{cao2017realtime}. Then, we follow HuMMan~\cite{cai2022humman} to derive 3D keypoints annotations $P_{3D}$ by triangulation with camera parameters obtained in the spatial alignment, in which we also import smoothness and bone length constraints for $P_{3D}$ to reduce temporal jitter and improve human shape consistency. 

\subsubsection{SMPL Fitting}
Multi-view SMPL fitting was a well-solved problem, with the inclusion of 3D joint, prior, smooth, and shape regularization errors.
However, due to the occlusion of facial and hand areas by the EgoSensorKit (HMD and controllers) and the Kinect camera's limited resolution, it's challenging to ensure the accuracy of the corresponding joint detection.
This results in unreasonable SMPL fitting for wrist and head joint rotations. To tackle this problem, we incorporate the global rotation of head $R_{head}$ and wrist joints $R_{wrist}$, which are transferred from the collected IMU rotations of hand controllers $R_{controller}$ and headset $R_{headset}$ by $R_{head}=R_{headset}^{head}\ R_{headset}$ and $R_{wrist}=R_{controller}^{wrist}\ R_{controller}$, where $R_{controller}^{wrist}$ and $R_{headset}^{head}$ are the constant transformation matrix obtained by statistical methods with a large amount of data collected in the standard sensor-wearing settings. Moreover, we leverage the calibrated leg motion tracker data, which represents the knee joint rotation $R_{knee}$, to constrain the lower leg pose. With the keypoint annotations and five joint rotations obtained above, we fit the SMPL parameters by minimizing the following energy function:
\begin{equation}
\label{eq:smpl_fitting}
\begin{split}
E(\theta, \beta)=  
&\lambda_{rot} E_{rot} + \lambda_{joint} E_{joint} + \lambda_{prior} E_{prior} + \\
& \lambda_{smooth} E_{smooth}  +\lambda_{reg} E_{reg},
\end{split}
\end{equation}
where $\theta \in \mathbb{R}^{75}$ and $\beta\in \mathbb{R}^{10}$ are optimized SMPL pose and shape parameters and $\lambda_{*}$ are balance weight.
For occluded joints, we introduce the rotation term to encourage the pose consistency with transferred IMU data as follows: 
\begin{equation}
E_{rot} = \sum_{j}||\mathcal{F}(\theta)_{j}-R_{j}||,
\end{equation}
where $j \in \{head, wrist, knee\}$ and $\mathcal{F}$ indicate a forward kinematic (FK) to get the joint global rotation. 
Other energy terms are like previous works \cite{bogo2016keep, cai2022humman, pavlakos2019expressive}, in which $E_{joint}$ minimizes the 3D distance between $P_{3D}$ and regressed SMPL joints. $ E_{prior}(\theta)$ is Vposer prior from SMPLify-X. $E_{smooth}$ helps to keep smooth pose tracking, while the shape regularization term $E_{reg}$ penalizes large shape variance.
With the space alignment result, the SMPL results could transfer from world space to the egocentric camera coordinate and obtain 2D pose annotations on egocentric images.

\section{A New Baseline Method: MEPoser}
To demonstrate the significance of the \dataname dataset and to inspire new designs for multimodal egocentric HPE, we introduce a new baseline method called \textbf{M}ulti-modal \textbf{E}gocentric \textbf{Pose} Estimato\textbf{r} (MEPoser). 
MEPoser takes multimodal inputs, including stereo egocentric images and inertial measurements, to extract multimodal representations and perform real-time HPE on a standalone HMD. 
As shown in Fig.~\ref{fig-baseline}, MEPoser consists of three components. (1) A multimodal fusion encoder extracts object representations at each frame from multimodal input data. (2) A temporal feature encoder composed of long short-term memory (LSTM) modules and feed-forward networks generates latent variables containing temporal information incorporated from past frames. (3) With the temporal aggregated multimodal features, two MLP-based (multi-layer perception) heads regress the pose and shape parameters of the SMPL model respectively.

\begin{table*}[h]
  \centering
  \begin{tabular}{l|l|ccccccc} 
    \toprule
    Dataset & Method & MPJRE$\downarrow$ & MPJPE$\downarrow$ & PA-MPJPE$\downarrow$ & UpperPE$\downarrow$ & LowerPE$\downarrow$ & RootPE$\downarrow$ & Jitter$\downarrow$\\
    \midrule
    \multirow{3}{*}{Protocol 1} & UnrealEgo & -& 5.5 & 3.9 & 4.0 & 7.7 & 4.2 & 592.5\\
    &HMD-Poser & 4.6  & 5.8 & 2.8 & 4.8 & 7.1 & 5.8 & \textbf{114.9}\\
    &MEPoser-CV & 5.4  & 4.5 & 2.9 & 3.3 & 6.3 & 3.8 & 511.0\\
    &MEPoser-IMU & 5.0  & 6.2 & 3.6 & 5.0 & 8.0 & 5.2 & 121.7\\
    &MEPoser-Full & \textbf{4.1}  & \textbf{3.7} & \textbf{2.5} & \textbf{2.7} & \textbf{5.1} & \textbf{3.2} & 161.8\\
    \midrule
    \multirow{3}{*}{Protocol 2} &UnrealEgo & -  & 6.4 & 4.3 & 4.6 & 8.9 & 5.0 & 610.5\\
    &HMD-Poser &  4.9 & 7.0 & 3.4 & 5.2 & 9.7 &7.2 & 165.7\\
    &MEPoser-CV & 6.0  & 5.4 & 3.5 & 3.8 & 7.8 & 4.4 & 566.5\\
    &MEPoser-IMU & 5.7  & 7.1 & 4.2 & 5.4 & 9.9 & 6.6 & \textbf{161.7}\\
    &MEPoser-Full &  \textbf{4.7} & \textbf{4.8} & \textbf{2.9} & \textbf{3.2} & \textbf{7.0} & \textbf{3.8} & 204.9\\
    \bottomrule
  \end{tabular}
\caption{Quantitative comparison between MEPoser and single-modal methods on our \dataname dataset}
\label{tab:method_compare}
\end{table*}

\subsection{Multimodal Fusion Encoder} The multimodal encoder first has separate feature encoders for different modalities, i.e., two weight-sharing CNN backbones for images and an MLP network for IMU data.
To make MEPoser run in real-time on HMD, we use a lightweight RegNetY-400MF~\cite{radosavovic2020designing} backbone, which takes stereo images $\{Img_{left}^{t_{i}}, Img_{right}^{t_{i}}\} \in \mathbb{R}^{640\times480\times1}$ as inputs, and generates 2D image features represented as $\{\mathrm{F}_{left}^{t_{i}}, \mathrm{F}_{right}^{t_{i}}\} \in \mathbb{R}^{80\times60\times256}$.
These features are then concatenated and forwarded to a few convolution layers to infer a set of heatmaps $\{\mathrm{H}_{left}^{t_{i}}, \mathrm{H}_{right}^{t_{i}}\} \in \mathbb{R}^{80\times60\times J}$. Here we predict 22 joints of the SMPL, i.e., $J=22$. 
To train the RegNetY-400MF backbone, we calculate the binary cross-entropy with logits loss (BCEWithLogitsLoss) $\mathcal{L}_{heatmap}^{2D}$ between the GT heatmaps and the estimated 2D heatmaps.
Then, the predicted heatmaps are flattened and forwarded to an MLP network to obtain the image feature.  We have obtained the IMU and image features so far. To boost the performance of the pose estimation, we added a 3D module to estimate the 3D joint positions in both the local camera coordinate and the global world coordinate. Specifically, given the image features from stereo heatmaps, an MLP network first encodes them to estimate the 3D joint positions in the local camera coordinate $\mathrm{\hat{P}}_{local}^{J\times3}$. Then, these joints are transferred to the global SMPL coordinate $\mathrm{\hat{P}}_{global}^{J\times3}$ with the offline calibration results and the online headset's 6DoF data. $\mathrm{\hat{P}}_{local}^{J\times3}$ and $\mathrm{\hat{P}}_{global}^{J\times3}$ are used to calculate the 3D joint loss $\mathcal{L}_{joints}^{Local}$ and $\mathcal{L}_{joints}^{Global}$, respectively.
Next, the joint positions $\mathrm{\hat{P}}_{global}^{J\times3}$ are flattened and forwarded to an MLP network to obtain the 3D joint features. 
Finally, the IMU, image, and 3D joint features are concatenated to output the multimodal fused feature $f^{t_{i}}$.

\subsubsection{Temporal Feature Encoder} 
As demonstrated in HMD-Poser~\cite{dai2024hmd}, temporal correlation information is the key to tracking accurate human motions. However, the multimodal fused features $\{f^{t_{i}}\}$ are still temporally isolated. To solve this problem, Transformer and RNN are adopted in existing methods. Although Transformer-based methods~\cite{zheng2023realistic} have achieved state-of-the-art results in HPE, their computational costs are much higher than those of RNN-based methods. To ensure our method runs in real-time on HMDs, we introduce a lightweight LSTM-based temporal feature encoder. Specifically, the encoder is composed of a stack of $N=3$ identical blocks. And each block has two sub-layers. The first is an LSTM module to learn the temporal representation, and the second is a simple fully connected feed-forward network. We employ a residual connection followed by layer normalization.

\subsection{SMPL Decoder} 
The SMPL decoder first adopts two regression heads to estimate the local pose parameters $\theta^{t_{i}}$ and the shape parameters $\beta^{t_{i}}$ of SMPL. Both regression heads are designed as a 2-layer MLP. Then, it uses an FK module to calculate all joint positions $\mathrm{\hat{P}}_{SMPL}^{J\times3}$ with $\theta^{t_{i}}$, $\beta^{t_{i}}$, and the online head's 6DoF data from headset.
We define the SMPL loss function $\mathcal{L}^{SMPL}$ as a combination of root orientation loss $\mathcal{L}_{ori}$, local pose loss $\mathcal{L}_{lrot}$, global pose loss $\mathcal{L}_{grot}$ and joint position loss $\mathcal{L}_{joint}$. All these losses are calculated as the mean of absolute errors (L1 norm) between the predicted results and the ground-truth values.

\subsection{Training MEPoser} 
For the overall training loss, we combine a smooth loss $\mathcal{L}_{smooth}$ with the above losses, including 2D heatmap loss $\mathcal{L}_{heatmap}^{2D}$, 3D joint loss $\mathcal{L}_{joints}^{Local}$, $\mathcal{L}_{joints}^{Global}$ and SMPL loss $\mathcal{L}_{SMPL}$. The smooth loss from HMD-Poser~\cite{dai2024hmd} is adopted to further enhance the temporal smoothness.
\begin{equation}
\label{eq:loss_total} 
\begin{split}
     \mathcal{L} = & \lambda_{hp} \mathcal{L}_{heatmap}^{2D} + \lambda_{ljoints} \mathcal{L}_{joints}^{Local} + \lambda_{gjoints} \mathcal{L}_{joints}^{Global} \\
    & + \lambda{smpl} \mathcal{L}_{SMPL} + \lambda_{smooth} \mathcal{L}_{smooth},
\end{split}
\end{equation}

\section{Experiment}

\begin{figure*}[!htb]
    \centering
    \includegraphics[width=0.98\linewidth]{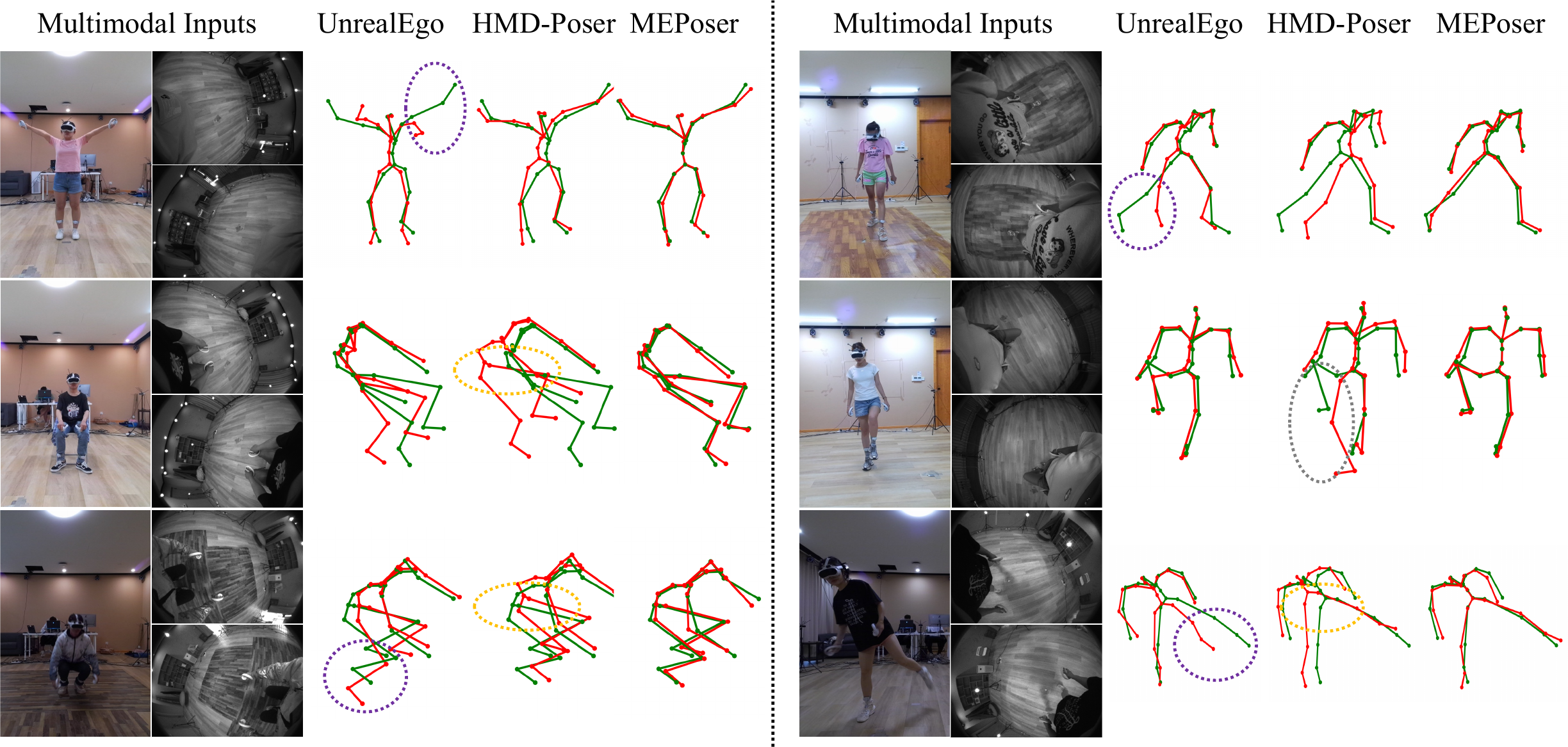}
    \caption{Qualitative comparison between ours and single-modal methods (GT: green, estimation: red).  Ours relieves joint invisibility in egocentric images (purple), IMU data drifting (yellow), and ambiguous measurements in slow motions(gray).}
    \label{fig-comparition}
\end{figure*}

\subsection{Dataset Splitting}
We split the dataset into three parts: one for training (70\%) and two separate testing sets based on different protocols, as follows. The training set comprises 615 sequences captured by 38 individuals, covering 20 daily actions involving upper-body, lower-body, and full-body movements. 
For testing, \textbf{Protocol 1} (16\%) contains 141 sequences with the same set of actions but performed by 8 different subjects than those in the training set, to evaluate cross-subject generalization. \textbf{Protocol 2} (14\%) is designed to assess the model's effectiveness and robustness in more general scenarios, consisting of 129 sequences involving 19 unseen actions and 20 unseen subjects not present in the training set.

\subsection{Comparison}

To validate the dataset and the corresponding baseline methods, we conducted comparisons using our dataset between MEPoser against the latest single-modal methods. The quantitative results in Tab. \ref{tab:method_compare} show that MEPoser outperforms existing single-modal methods for egocentric HPE. Compared to Unrealego~\cite{akada2022unrealego} which takes stereo egocentric images as inputs, our method reduces the MPJPE (Mean Per Joint Position Error, cm) by 32.7\% and 25\% in protocol1 and protocol2, respectively. Notably, MEPoser significantly enhances the smoothness of estimation results by using a temporal LSTM structure. Besides, our method obtains 27.5\% and 23.8\% enhancement in joint location precious in contrast to EgoPoseFormer~\cite{yang2024egoposeformer}. In comparison with whole-body IMU-based method HMD-Poser~\cite{dai2024hmd}, MEPoser shows 36.2\% and 31.4\% reduction in MPJPE on two test sets by combining the egocentric image features and slight improvement of joint rotation accuracy according to the MPJRE (Mean Per-Joint Rotation Error, $^\circ$). The results also demonstrate the improved generalizability of MEPoser across subjects and actions, validating the effectiveness of the multimodal setting and the value of our dataset in solving egocentric HPE. We also include ablation experiments to investigate the impact of different components in our network and the results demonstrate that each modality contributes to performance improvements. 

Qualitative comparisons are shown in Fig. \ref{fig-comparition} using various test sequences, featuring different actions and environments. MEPoser could relieve the limitations of the single-modal methods by exploiting the complementary between the vision and IMU signals. Specifically, our method could deal with issues like self-occlusion and out-of-FOV problems in egocentric images by utilizing IMU features and temporal information to get more accurate pose results. Additionally, our approach mitigates the sparsity, drifting, and ambiguous measurements in slow motions of IMU signals by incorporating visible body joints in egocentric images. 

\subsection{Annotation Cross-validation}

\begin{table}
  \centering
  \begin{tabular}{cc|cc}
    \toprule
    Action & MPJPE & Action & MPJPE\\
    \midrule
    walking & 2.61 & kicking & 2.12 \\
    hand-waving &  2.72 & lunge &2.23 \\
    taichi & 2.98 & boxing & 2.73 \\
    shuttlecock-kicking & 2.65 &
     dancing & 2.49\\
    marching-in-place & 2.08 & \textbf{Average} & \textbf{2.51}\\
    \bottomrule
  \end{tabular}
\caption{Cross-validation for dataset annotations }
\label{tab:evaluation}
\end{table}

To further assess the accuracy of our dataset annotations, we randomly capture 9 motion sequences for cross-validation by simultaneously using both our system and an optical marker-based motion capture system. 
The subject wears a suit attached with reflective markers tracked by the OptiTrack system. Marker-based SMPL parameters are then derived from the MoCap data using MoSh++~\cite{loper2014mosh}, with a temporal filter applied to reduce jittering.
As shown in Tab. \ref{tab:evaluation}, the low error metrics and variance demonstrate the robustness of our annotation pipeline and the high quality of the dataset. Besides the validation, every sequence of our dataset has been inspected manually to eliminate the data with erroneous annotations.

\section{Conclusion}
In this paper, we introduce \dataname, a novel multimodal human motion dataset designed for egocentric HPE. It includes synchronized egocentric images and IMU signals from a real VR product suite, with SMPL annotations in the same world coordinate system. To enhance generalization in real-world applications, we collected a diverse range of data across various actions and individuals.  We also present MEPoser, a new baseline HPE method that combines image and IMU inputs for real-time HPE on a standalone HMD. MEPoser effectively demonstrates the benefits of multimodal fusion, improving accuracy and addressing the limitations of previous single-modal methods. This approach serves as an initial exploration, inviting further research of egocentric HPE with multimodal data. We believe releasing this dataset and method will accelerate the practical implementation of HPE with body-worn sensors in future VR/AR products.

\newpage
\bibliography{aaai25}

\end{document}